\documentclass[conference]{IEEEtran}
\ifCLASSINFOpdf
\else
\fi
\begin{document}
%
\title{\huge{Machine Learning Techniques in Cognitive Radio Networks}}

\author{Peter Hossain$^1$, Netherlands\\
Adaulfo Komisarczuk$^2$, Garin Pawetczak$^2$,  Germany\\
Sarah Van Dijk$^3$,  France\\
Isabella Axelsen$^4$, Denmark\\}


%


\maketitle

\begin{abstract}
Cognitive radio is an intelligent radio that can be programmed and configured dynamically to fully use the frequency resources that are not used by licensed users. It defines the radio devices that are capable of learning and adapting to their transmission to the external radio environment, which means it has some kind of intelligence for monitoring the radio environment, learning the environment and make smart decisions. In this paper, we are reviewing some examples of the usage of machine learning techniques in cognitive radio networks for implementing the intelligent radio.
\end{abstract}


%
\IEEEpeerreviewmaketitle

\section{Introduction}
\indent The concept of cognitive radio was first proposed by Joseph Mitola III. It was designed for mitigating the scarcity problem of limited radio spectrum by improving the utilization of the spectrum. Based on Joseph's proposal, it was a novel approach in wireless communications, which was described as the point in which wireless personal digital assistants (PDAs) and the related networks are sufficiently computationally intelligent about radio resources and related computer-to-computer communications to detect user communications needs, and to provide radio resources and wireless services most appropriate to those needs \cite{cr1}\cite{cr2}. \\
\indent Cognitive radio is considered as a goal towards which a software-defined radio platform should evolve: a fully reconfigurable wireless transceiver which automatically adapts its communication parameters to radio environment and user demands. Traditional regulatory structures have been built for an analog model and are not optimized for cognitive radio. Regulatory bodies in the world as well as different independent measurement campaigns found that most radio frequency spectrum was inefficiently utilized. The frequency bands that have been allocated to different licensed user are insufficiently utilized. To address the frequency shortage issue, cognitive radio is treated as a promising solution.\cite{cr2}\\
\indent To perform its cognitive tasks, the CR users should have the ability to monitor the surrounding radio environment, take essential measurements and make intelligent decisions to reuse of some under untilized frequency resources. There are several main areas of study as show following:\\
\indent Spectrum sensing: Detecting unused spectrum and allow the CR user to access it without bringing harmful interference to the licensed users. It is an important requirement of the cognitive radio network to sense empty spectrum.\\
\indent Spectrum management: Capturing the best available spectrum to meet user communication requirements while not creating harmful interference to the licensed users. It also responsible for monitor how long the
spectrum holes are available for the CR user to access them.\\
\indent  Spectrum sharing: it takes care of distributing the sensed idle spectrum bands among the CR users and meet certain level of fairness among all CR users \cite{share}.\\
\indent Machine learning is an approach of implementing Artificial Intelligence (AI), which is defined as the intelligence that exhibited by machines or software, dealing with the construction and study of systems that can learn from data, rather than follow only explicitly programmed instructions \cite{AI}. Since cognitive radio is defined as an intelligent radio for make the best use of the frequency resources, machine learning  have been introduced to CR for implementing the intelligent radio \cite{brain1}. In this paper, we study the current application of machine learning in cognitive radio in variety of areas.

\section{Machine Learning for Spectrum Sensing}
\indent Spectrum sensing is required in cognitive radio in order to help the CR users find the spectrum holes. In \cite{sensing1}, the author proposed a cooperative spectrum sensing (CSS) schemes based on
machine learning techniques. In the context
of CSS, they treat an ¡°energy vector¡±, each component of
which is an energy level estimated at each CR device, as
a feature vector. Then, the classifier categorizes the energy
vector into one of two classes: the ¡°channel available class¡±
(corresponding to the case that no PU is active) and the ¡°channel unavailable class¡± (corresponding to the case that at least
one PU is active). The proposed machine learning-based CSS techniques have
the following advantages over the traditional CSS techniques. 1. The proposed techniques are capable of implicitly learn-
ing the surrounding environment (e.g., the topology of
the PU and the CR networks and the channel fading) in
an online fashion. Therefore, the proposed techniques are
much more adaptive than the traditional CSS techniques,
which need prior knowledge about the environment for
optimization. 2. The proposed techniques can describe more optimized
decision region
1
on the feature space than the traditional
CSS techniques (e.g., OR/AND-rule-based and linear
fusion techniques) can, which results in better detection
performance.

\indent In \cite{sensing2},  a machine learning based multiband spectrum sensing
policy has been proposed. In the policy the greedy method has
been used to track the occupancy statistics of the primary user and to
estimate the detection performance of the secondary users. Employing the greedy method the policy is able to select the subbands to
be sensed and possibly accessed that consistently provide spectrum
opportunities with high throughput for the secondary network, thus
increasing the overall sum data rate. Using the knowledge about the
detection performance, the policy assigns those secondary users to sense the interesting subbands that are able to meet a desired miss
detection probability. The sensing assignment is found by formulating a binary integer programming problem where the objective
is to minimize the number of sensors
D
per subband, while ensuring the desired detection performance at each subbands. The benefit of minimizing the number of sensors per subband is obviously
improved energy efficiency. Instead of reducing the SU battery consumption, this type of policy could also be used to simultaneously scan more subbands by employing the unassigned secondary users
to sense other parts of the spectrum.
\hfill mds

\section{Machine Learning for Spectrum Sensing}
\indent Spectrum sensing is required in cognitive radio in order to help the CR users find the spectrum holes. In \cite{sensing1}, the author proposed a cooperative spectrum sensing (CSS) schemes based on
machine learning techniques. In the context
of CSS, they treat an ¡°energy vector¡±, each component of
which is an energy level estimated at each CR device, as
a feature vector. Then, the classifier categorizes the energy
vector into one of two classes: the ¡°channel available class¡±
(corresponding to the case that no PU is active) and the ¡°channel unavailable class¡± (corresponding to the case that at least
one PU is active). The proposed machine learning-based CSS techniques have
the following advantages over the traditional CSS techniques. 1. The proposed techniques are capable of implicitly learn-
ing the surrounding environment (e.g., the topology of
the PU and the CR networks and the channel fading) in
an online fashion. Therefore, the proposed techniques are
much more adaptive than the traditional CSS techniques,
which need prior knowledge about the environment for
optimization. 2. The proposed techniques can describe more optimized
decision region
1
on the feature space than the traditional
CSS techniques (e.g., OR/AND-rule-based and linear
fusion techniques) can, which results in better detection
performance.

\indent In \cite{sensing2},  a machine learning based multiband spectrum sensing
policy has been proposed. In the policy the greedy method has
been used to track the occupancy statistics of the primary user and to
estimate the detection performance of the secondary users. Employing the greedy method the policy is able to select the subbands to
be sensed and possibly accessed that consistently provide spectrum
opportunities with high throughput for the secondary network, thus
increasing the overall sum data rate. Using the knowledge about the
detection performance, the policy assigns those secondary users to sense the interesting subbands that are able to meet a desired miss
detection probability. The sensing assignment is found by formulating a binary integer programming problem where the objective
is to minimize the number of sensors
D
per subband, while ensuring the desired detection performance at each subbands. The benefit of minimizing the number of sensors per subband is obviously
improved energy efficiency. Instead of reducing the SU battery consumption, this type of policy could also be used to simultaneously scan more subbands by employing the unassigned secondary users
to sense other parts of the spectrum.

\section{Machine Learning for Power Allocation}
\indent In \cite{power1}, the authors have proposed a decentralized Q-learning
algorithm to solve the problem of power allocation in a
secondary network made up of several independent cells, given
strict limit for the allowed aggregated interference on the
primary network. It compared the performance of the
decentralized algorithm with the performance of the optimal
centralized power allocation algorithm.
It has been showed that the implementation of a cost func-
tion that penalizes the actions leading to a higher than required
secondary SINR gives better results than the implementation
of a cost function without such penalty.
The Q-learning algorithm has been showed to converge
faster when its frequency of execution increases, untill the
frequency reaches an upper bound where the increase of the
convergence speed gets insignificant.
Finally, it has been observed that the strategy of keeping
the exploration parameter constant in the learning algorithm
is less efficient than using a linearly decreasing parameter or
implementing an alternance between full exploration and full
exploitation, this latest exploration policy leading to the fastest
convergence of the algorithm.

\section{Machine Learning for Radio Access Technology}
\indent In \cite{RA1}, it presents an experimental testbed for radio
access technology (RAT) recognition in cognitive radio
networks, where there could be concurrent transmissions from
Primary Users (PUs) owning a frequency band usage license
and Secondary Users (SUs) who operate on a license-exempt, basis. Primary and secondary users¡¯ signal recognition by SUs
will help in accommodating the SUs' transmissions adequately
¨C which might be considered as interference to the PUs
otherwise. In \cite{3a}, the authors have studied the classification of
many RATs e.g. GSM, UMTS, DECT, DAB, etc. The
recognition was based on the a priori knowledge of these
RATs¡¯ channel bandwidths. This
approach will have a low
classification performance, since most of the current RATs use
scalable channel bandwidths, e.g., Scalable-OFDMA, and thus
one RAT could be mistakenly classi
fied for another. In \cite{4b}, an
automatic network recognition method was devised for the
classification of WiFi and Bluetooth transmissions. However,
these two radio communication protocols use inherently
different multiplexing techniques, namely, Direct Sequence
Spread Spectrum (DSSS) and Frequency Hopping Spread
Spectrum (FHSS); where the former is using a fixed and wider
channel bandwidth than the latter, and the latter is using
pseudo-randomly hopping transmissions of very short duration,
i.e., 625¦Ìsec. Furthermore, Bluetooth is not as versatile as other
RATs in handling high data rates, and thus its use in sub-GHz
shared spectrum bands is not predicted for the near future.

\section{Machine Learning for Signal Classification}
\indent \cite{signal1} present a study of multi-class
signal classification based on automatic modulation recognition
through Support Vectrom Machines (SVM). It implemented
a simulated model of an SVM signal classifier trained to recognize seven distinct modulation schemes;
five digital (BPSK, QPSK, GMSK, 16-QAM and 64QAM) and two analog ones (FM and
AM). The signals are generated using realistic carrier frequency,
sampling frequency and symbol rate values, and pulse-shaping
filters types. This paper also reports on the on-going experimental work
by using software defined radios (SDR) to implement this as a
part of our cognitive radio network testbed. The results show
that overall performance of classifier is very good; the trained
SVM correctly classifies signals with
85 to 98\%
probability in its current state of development.
\indent In \cite{signal2}, it presents two approaches to the problem
of automatic modulation classification in cognitive radios. The
first approach deals with classifying distinct QAM modulations.
The proposed solution consists of a SVM classifier that uses
as input parameters the distance between the received symbol
and its nearest neighbor in each constellation. The results
obtained by the proposed classifier is compared to a state-of-art
technique. The comparison show a good performance for the
16-QAM, 32-QAM and 64-QAM modulation schemes. In the
second approach, the cyclic spectral analysis is adopted as the
method for extracting the signals features, and the SVM classifier
is used for the pattern recognition stage. The proposed method
achieved excellent results for the AM, BPSK, QPSK and BFSK
modulation schemes. In both approaches, different scenarios are
evaluated which may serve as a starting point for researchers
who want to compare results systematically.

\section{Machine Learning for Medium Access Protocol Identification}
\indent In \cite{MAC1}, the author investigate the medium access control protocol identification for applications in
cognitive MAC. MAC protocol identification
enables CR users to sense and identify the
MAC protocol types of any existing transmissions (primary or secondary users). The identification results will be used by CR users to
adaptively change their transmission parameters in order to improve spectrum utilization,
as well as to minimize potential interference to
primary and other secondary users. MAC protocol identification also facilitates the implementation of communications among
heterogeneous CR networks. In this article, the authors
consider four MAC protocols, including
TDMA, CSMA/CA, pure ALOHA, and slotted
ALOHA, and propose a MAC identification
method based on machine learning techniques.
Computer simulations are performed to evaluate the MAC identification performance.

\section{Machine Learning for Attack Detection in Cognitive Radio Networks}
\indent In \cite{attck1},  a SVM algorithm for jamming detection at
the base station is proposed. The proposed jamming detection
algorithm is modeled in a TETRA simulated base station
receiver. It is shown that using SVM algorithm improves the
performance of the jamming detection algorithm and
decreases the hardware complexity. The application
of the
proposed jamming detection algorithm is simulated a
nd
compared with other conventional methods and simulation
results exhibit a noticeable improvement in presence of a wide
range of user terminal velocities and different jammers. The
proposed jamming detection enhancements can be used
in to a
base station of any wireless communication systems
as it has
been applied in the TETRA base station receiver. The
proposed algorithm for jamming detection is independent from
the operation and architecture of the wireless communication
system used and they are described only to a degree
that assist
in comprehending the work. An example of a wireless
communication system wherein the proposed algorithm
can be
applied is a TETRA system, although it can be extended to
other systems such as GSM Global System for Mobile
Communications and UMTS Universal Mobile
Telecommunications System.

\section{Conclusions}
Cognitive radio is an intelligent radio when can made learn from the radio environment and intelligent decisions to use the underutilized frequency resources. In this paper, we reviewed the serval typical application scenarios based on machine learning techniques in cognitive radio. It shows the machine learning is a very promissing for implement intelligent cognitive radio.

\end{document}